\documentclass{article}
\PassOptionsToPackage{numbers, compress}{natbib}
\usepackage[preprint]{neurips_data_2023}
\usepackage[utf8]{inputenc} % allow utf-8 input
\usepackage{hyperref}       % hyperlinks
\usepackage{url}            % simple URL typesetting
\usepackage{booktabs}       % professional-quality tables
\usepackage{amsfonts}       % blackboard math symbols
\usepackage{nicefrac}       % compact symbols for 1/2, etc.
\usepackage{microtype}      % microtypography
\usepackage{xcolor}         % colors

\usepackage{times}
\usepackage{graphicx}
\usepackage{CJKutf8}
\usepackage{tabularx}
\usepackage{float}
\usepackage{subfigure}
\usepackage{caption}
\usepackage{longtable}
\usepackage{supertabular}
\usepackage{arydshln} 
\usepackage{rotating}
\usepackage{enumitem}
\usepackage{amsmath}

\usepackage{epsfig}
\usepackage{amssymb}
\usepackage{algorithm}
\usepackage{algpseudocode}
\usepackage{threeparttable}
\usepackage{booktabs}
\usepackage{mathrsfs}
\usepackage{bbding}

\usepackage{multirow}

\setenumerate[1]{itemsep=0pt,partopsep=0pt,parsep=\parskip,topsep=5pt}
\setitemize[1]{itemsep=0pt,partopsep=0pt,parsep=\parskip,topsep=5pt}
\setdescription{itemsep=0pt,partopsep=0pt,parsep=\parskip,topsep=5pt}

\title{Tachikuma: Understading Complex Interactions with Multi-Character and Novel Objects by Large Language Models}

\author{\small Yuanzhi Liang$~\textsuperscript{\rm 1}$, Linchao Zhu$~\textsuperscript{\rm 2}$, Yi Yang$~\textsuperscript{\rm 2}$    \\ 
        \small{$~\textsuperscript{\rm 1}$ University of Technology Sydney}, 
	\small{$~\textsuperscript{\rm 2}$ Zhejiang University}
	\\{\tt\small {yuanzhi.Liang}@student.uts.edu.au} \vspace{-0.15cm}
        \\{\tt\small {zhulinchao7}@gmail.com} \vspace{-0.15cm}
        \\{\tt\small {yangyics}@zju.edu.cn}	
}
\date{}

\begin{document}

\maketitle

\begin{abstract}
Recent advancements in natural language and Large Language Models (LLMs) have enabled AI agents to simulate human-like interactions within virtual worlds. However, these interactions still face limitations in complexity and flexibility, particularly in scenarios involving multiple characters and novel objects. Pre-defining all interactable objects in the agent's world model presents challenges, and conveying implicit intentions to multiple characters through complex interactions remains difficult. To address these issues, we propose integrating virtual Game Masters (GMs) into the agent's world model, drawing inspiration from Tabletop Role-Playing Games (TRPGs). GMs play a crucial role in overseeing information, estimating players' intentions, providing environment descriptions, and offering feedback, compensating for current world model deficiencies. To facilitate future explorations for complex interactions, we introduce a benchmark named Tachikuma, comprising a Multiple character and novel Object based interaction Estimation (MOE) task and a supporting dataset. MOE challenges models to understand characters' intentions and accurately determine their actions within intricate contexts involving multi-character and novel object interactions. Besides, the dataset captures log data from real-time communications during gameplay, providing diverse, grounded, and complex interactions for further explorations. Finally, we present a simple prompting baseline and evaluate its performance, demonstrating its effectiveness in enhancing interaction understanding. We hope that our dataset and task will inspire further research in complex interactions with natural language, fostering the development of more advanced AI agents.
\end{abstract}

\section{Introduction}
%\noindent\emph{A child use such primitive forms of language when he learns to task. Here the teaching of language is not explaining, but training.}
\noindent\emph{... the teaching of language is not explaining, but training.}

-- Ludwig Josef Johann Wittgenstei, Philosophical Investigations 

\iffalse 
Format Hardback | 592 pages
Dimensions 160 x 235 x 35mm | 972g
Publication date 13 Nov 2009
Publisher John Wiley and Sons Ltd
Imprint Wiley-Blackwell (an imprint of John Wiley & Sons Ltd)
Publication City/Country Chicester, United Kingdom
Language English
Edition Statement 4th Edition
ISBN10 1405159286
ISBN13 9781405159289
page 7 
\fi

In recent years, there has been a growing interest in constructing AI agents capable of simulating and supporting human-like interactions across various domains. Notably, some agents have exhibited exceptional performance, surpassing human abilities in games like MOBA, Starcraft, poker, and Go. Building on the advancements in Large Language Models (LLMs), researchers have extended agent interactions to incorporate natural language. For instance, Park et al. \cite{park2023generative} have introduced generative agents that engage in free-form interactions using natural language, thereby creating virtual worlds where agents reside and even demonstrate spontaneous activities such as hosting parties. Similarly, Liu et al. \cite{liu2023training} have developed simulated societies in which LLM-powered agents engage in the virtual world and can support some discussions for social problems. These recent developments hold promise for advancing AI agents by leveraging natural language as an interactive tool, enabling them to exhibit more human-like behaviors. Furthermore, the exploration of phenomena resulting from endowing agents with more powerful language abilities for interaction can offer valuable insights. As discussed in the philosophical investigation, Ludwig Josef Johann Wittgenstein emphasized that teaching language is a form of training rather than mere explanation. General human communication is similar to engaging in a language game. Language serves as a fundamental tool for human interaction with the environment, facilitating the transmission of information, communication, negotiation, and cooperation within human groups, and contributing to the overall functioning of society. While the relationship between language and intelligence remains an open question, it is always worth exploring the potential evolution of more powerful and autonomous agents that can interact using natural language.

Going further with agent interactions, we have yet to fully empower the sufficient openness and freedom in the interactions between agents and the world. Existing approaches have often imposed constraints on agent interactions, leading to limited complexity and diversity in their capabilities. These constraints arise from the lack of interactions involving novel objects and multiple characters. While some prior research has explored language-based interaction abilities in generative agents~\cite{park2023generative}, their diversity remains restricted, focusing on a limited range of interactable objects. Additionally, previous works have primarily concentrated on two-character communication without considering implicit intentions through complex interactions. Such interactions fail to encompass nuanced behaviors (e.g., refusal, persuasion, group decision making, coalition building), akin to real-time communications involving multi-characters.

%Specifically, although some prior research has explored language-based interaction abilities in generative agents~\cite{park2023generative}, the diversity of these agents is still restricted, with a limited range of interactable objects considered. Previous works have primarily focused on two-character communication without the consideration of implicit intentions through complex interactions. Two agents present plain and simple interactions, which do not consider convey complex or implicit intentions like refusing, persuading, group decision making, coalition building, etc., close to the real-time communications with multi-characters. 

%failing to capture the complexity and realism of long and intricate multi-character interactions that are characteristic of real-time communication. 
%Furthermore, before we can develop better agent interactions, it is crucial to ensure that the agents first understand complex interactions. At present, we have not achieved the point where agents can autonomously engage in complex interactions themselves. Therefore, addressing the understanding challenge is a vital step towards empowering AI agents for more sophisticated and diverse interactions.
%Moreover, before empowering AI agents to engage in more complex and diverse interactions, we should first let the agents understand the complex interactions close to the real-human, which contains interactions with unlimited objects and characters. 

To address this challenge, we draw inspiration from tabletop role-playing games (TRPGs) and introduce a Game Master (GM) role into the agent's world model. TRPGs inherently offer highly complex and diverse interactions through natural language, involving multiple players in intricate and grounded multi-character scenarios. The GM oversees the game, provides scenario details, understands characters' intentions, and offers feedback on player actions, aligning with the requirements for a more comprehensive world model. Constructing and introducing a virtual GM capable of handling complex interactions with real humans could significantly enhance the feedback given to agents.
However, existing benchmarks in TRPG-related research lack the scope needed to develop a virtual GM that compensates for world model deficiencies. Current virtual GM works only explore short and simple interactions in limited rounds, lacking sufficient complexity. For instance, previous works have been derived from play-by-post forums~\cite{martin2018dungeons,callison-burch-etal-2022-dungeons}, where players contribute by writing and posting their responses on the forum. While, this asynchronous online communication introduces significant delays, with players often waiting for hours or even weeks to receive responses. As a result, data collected from such forums struggle to capture the vibrant and nuanced grounded semantics characteristic of real-time human interactions.
Moreover, the forum-based communication format tends to encourage players to respond to the immediate turn and provide formal written replies, thereby limiting the richness and groundedness of expressions that can be observed in real-time interactions with multi-characters. Consequently, previous works derived from forum data do not fully represent the diversity and complexity found in real-world multi-character interactions. More comprehensive and realistic benchmarks are needed to support the development of effective virtual GMs and address the deficiencies in agent world models.

In this paper, we take the first step towards enhancing the world model for agents by integrating a virtual GM role capable of handling complex real-time interactions with multiple characters. We propose a benchmark, named Tachikuma, designed to encourage the designation of the virtual GM to effectively handle these complex interactions, infer characters' intentions, and provide accurate feedback to corresponding characters.
Our benchmark consists of two components: a Multiple character and novel Object based interaction Estimation (MOE) task and a supporting dataset. In MOE, models are presented with intricate contexts extracted from TRPG log data, capturing real-time communications during gameplay. The objective is to infer character intentions and identify corresponding interactions, typically represented as skill checks, judged by a GM. The dataset supports the MOE task by providing long and intricate contexts from game logs, featuring interactions among multiple characters. The complexity of interactions among multiple characters, grounded in natural language, makes MOE a valuable testbed for evaluating abilities of virtual GMs.

Furthermore, in our dataset, we collect complex and long contexts with diverse real-human interactions from the game logs. Our dataset differs from conventional play-by-post forum data collection methods. Instead, we utilize data extracted from a Chinese TRPG forum\footnote{www.goddessfantasy.net}. These forum records, compiled by GMs after game ending, consist of voice recordings or real-time chat logs. This data source overcomes the limitations of play-by-post data collection, enabling us to extract long contexts with complex semantics similar to the real interactions. As these logs capture immediate communications, the interactions also exhibit higher groundedness, resulting in more vibrant and realistic responses akin to everyday conversations, as demonstrated in Fig.~\ref{fig:intro_moe}.
Moreover, our dataset encompasses not only the popular DND rules~\cite{gygax1974dungeons} but also a wide range of diverse game rules, including Call of Cthulhu (COC)~\cite{lovecraft2016call}, Pathfinder2 (PF2)~\cite{bulmahn2010pathfinder}, Savage Worlds (SW)~\cite{hensley2008savage}, etc. This diversity enhances the complexity and variety of our dataset. Building upon this dataset, we introduce MOE task, which consists of 1,003 context sections extracted from the game logs. Each section represents a complete adventure with multiple turns, showcasing intricate semantics. As shown in Tab.~\ref{tab.dataset_sta}, MOE includes an average of 32.12 turns per context excerpt, in contrast to previous works that typically involve only one turn.
The number of possible answers for characters and skills varies depending on the context, ranging from one to eleven. Additionally, specific game rules necessitate different skill categories for answers. For instance, considering the DND rule, there are 51 potential skills. These factors collectively contribute to MOE representing a challenging task for AI agents. The agent must demonstrate a comprehensive understanding of both the complex interactions, emulating human-like comprehension.
To provide a comprehensive assessment, we report the F-score as the final metric, separately for the predicted characters and overall intention answers. Evaluating character predictions reflects the accuracy of methods in inferring players' intentions. Simultaneously, evaluating overall answers offers insights into the understanding ability of both character intentions and the corresponding interactions.

Finally, we present a three-step prompting baseline for constructing an agent capable of handling interactions like a real-human GM in TRPGs. Our simple baseline serves to demonstrate the value of our task and dataset in understanding complex interactions. Our method incorporates prompts specifically related to existing characters, their intentions, and the associated skill checks. By utilizing these prompts, we guide LLMs in gradually comprehending the intricate interactions that occur between players. We thoroughly evaluate our baseline method and compare its performance with other prompting methods utilizing various LLMs within MOE task. 
The experimental results indicate that MOE task is solvable but still possesses a large room for further improvement.
Furthermore, leveraging the answers obtained from MOE task, we employ LLMs to generate responses that simulate a real-human GM in the games. To evaluate the quality of these generated responses, we invite numerous volunteers to provide subjective evaluations. The experimental results demonstrate that incorporating the improved understanding ability of the agent leads to higher levels of factual correctness, naturalness, and groundedness in the generated responses, closely resembling real-human interactions. These results further underscore the significance of understanding ability in constructing proficient agents and highlight the importance of our benchmark.
We hope our dataset and benchmark as valuable resources that will inspire the research community to delve into the understanding of complex interactions and contribute to the development of more capable AI agents.

Our contributions can be summarized as follows:

1. We introduce a Multiple character and novel Object based interaction Estimation (MOE) task, specifically addressing challenges in handling complex interaction like a real-human GM. This task serves as a valuable testbed for evaluating the abilities of constructing virtual GMs and contributes to advancements in developing more realistic agents.

%We present a Multiple character and novel Object based interaction Estimation (MOE) task, which is designed to tackle the challenges associated with complex multi-character interaction understanding. This task serves as a valuable testbed for evaluating the agents' understanding abilities and contributes to the ongoing advancements in constructing more vivid agents.

%1. We introduce Multiple character and Open instances based interaction Estimation (MOE) task, specifically designed to address the challenges of complex multi-character interaction understanding. This task serves as a valuable testbed for assessing the understanding abilities of agents, which is valuable in further improvements of world model designation for agents' interactions.

2. We collect a dataset for MOE to address the limitations in exploring long contexts and intricate multi-character interactions in real-time communications. This dataset bridges a crucial gap in the current research, offering a comprehensive resource for analyzing and understanding these complex interactions.

%We introduce a Multiple character and novel Object based Interaction dataset (MOD) to overcome the existing limitations in investigating long contexts and intricate multi-character interactions within real-time communications. This dataset fills a crucial gap in the current research, providing a comprehensive resource for analyzing and understanding these complex interactions.

%2. We create a new dataset, Multiple character based Open Interaction dataset (MOD), to address the limitations in exploring long contexts and complex multi-character interactions in real-time communications. This dataset fills a gap in the current research by providing a comprehensive resource for analyzing such complex interactions.

3. We introduce a prompting baseline and conduct a comprehensive evaluation of different prompting methods using a range of Large Language Models (LLMs) within MOE task. The experimental results indicate that MOE task is solvable, yet there is ample room for further improvement. %These results highlight the potential for improving the performance of agents in complex multi-character interactions within MOE task.

4. We conduct subjective evaluations based on the answers obtained from MOE. These evaluations show that better performances in MOE lead to higher levels of factual correctness, naturalness, and groundedness in the generated responses, which are crucial factors for creating a vivid agents. These results further underscore the significance of our dataset and task in improving AI agents.

\section{Related Work}
%Tabletop Role-Playing Games (TRPGs) are a form of interactive, narrative-driven game in which players assume the roles of characters in a fictional setting. This genre of games includes popular titles such as "Dungeons \& Dragons," "Pathfinder," and "Call of Cthulhu."
%In a typical TRPG, one player takes on the role of the game master (GM) or dungeon master (DM), who is responsible for creating the game world, narrating the story, and controlling non-player characters and events. The other players each control a single character and interact with the game world and the narrative presented by the GM.
%The outcomes of character actions in TRPGs are usually determined by a system of rules and dictated by dice rolls, although different games may utilize different rule sets and types of dice. The goal of these games is not necessarily to "win" in the traditional sense, but rather to participate in a collaborative storytelling experience and develop the characters and the story in interesting ways.
%TRPGs can be incredibly diverse, taking place in a variety of settings, from medieval fantasy worlds, to futuristic science-fiction universes, to modern-day detective mysteries, and anything in between. The flexibility and creativity inherent in TRPGs have led to their enduring popularity.

%\textbf{trpg game as an nlp challenge }

Tabletop Role-Playing Games (TRPGs) are immersive games where players assume different character roles in fictional settings, guided by a Game Master (GM) who provides relevant information to progress the game. These games involve diverse and complex grounded natural language interactions among multiple characters with distinct personalities and backgrounds. Due to the diversity and complexity, TRPGs serve as valuable testbeds~\cite{weir2022ontologically,louis2018deep,callison-burch-etal-2022-dungeons} for research in Natural Language Processing (NLP). Several works have explored NLP problems using TRPG game records. For instance, Louis et al.~\cite{louis2018deep} proposed predicting character actions based on previous interactions. Other works~\cite{si-etal-2021-telling,newman-liu-2022-generating} focused on generating flexible dialogue or descriptions in accordance with varying contexts or specific rules in TRPGs.

Furthermore, recent studies have commonly utilized play-by-post data from popular DND forums, providing a substantial corpus for research. This play-by-post format allows players to interact by posting replies, reducing participation barriers and generating a significant number of game rounds on the forum. Chris et al.~\cite{callison-burch-etal-2022-dungeons} have collected extensive corpus from these forums, resulting in the creation of TRPG dialogue datasets. Subsequently, Pei et al.~\cite{gandalf} filtered the dataset and developed a guidance generation task called GANDALF. Given the context from a single round, GANDALF predicts the guidance provided by the DM under the DND rule. Zhu et al.~\cite{zhu2023fireball} further extended the approach by constructing a more comprehensive and larger dataset using the play-by-post format in Discord, a messaging program. This dataset, named FIREBALL, contains additional game details such as dialogues, states, combat procedures, etc. It serves as a versatile testbed for language generation, particularly focusing on generating commands for games, including combat actions, checks, and dice rolls.

In this paper, we address the limitations of previous works in exploring more complex interactions. We introduce Multiple character and novel Object based interaction Estimation (MOE) task and Multiple character and a supporting dataset as valuable resources for interaction understanding for agents. Unlike previous approaches that rely on play-by-post formats, our dataset leverages game logs obtained from real-time interactions, providing a more grounded and complex semantics. MOE requires methods to answer questions about next acting characters and their corresponding actions. This task and dataset open up new possibilities for improving the agents with enhanced factual correctness, naturalness, and groundedness.

\section{Multiple character and novel Object based interaction Estimation}

\begin{figure*}[t]
    \centering
    \includegraphics[width=0.9\linewidth]{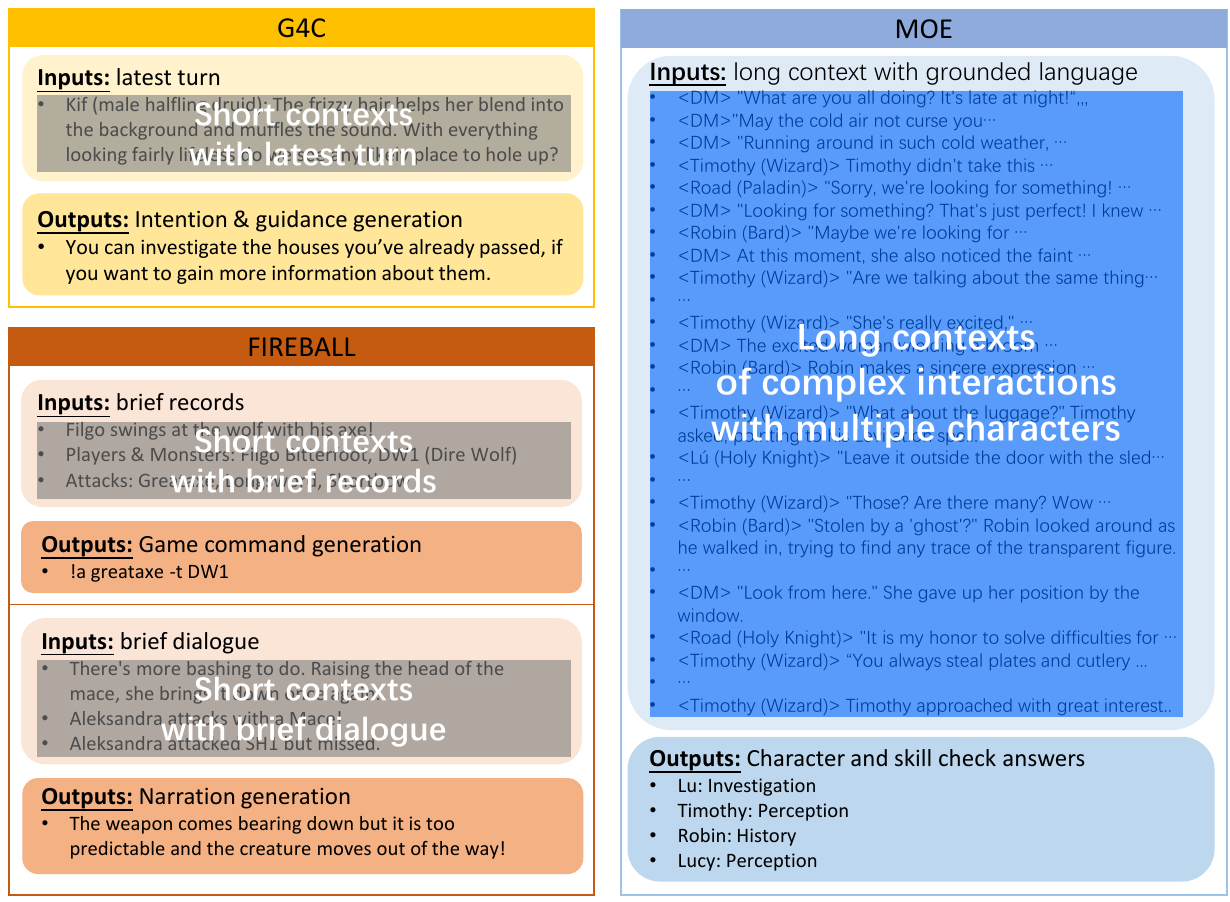}
    \caption{Examples of different tasks and datasets based on game logs of TRPG. Our MOE and MOD focuses on the understanding of long and complex interactions with Long contexts. }
    \label{fig:intro}
\end{figure*}

\subsection{Task Overview}
%We propose a novel task called  Complex Contexts based intention Answering (C2A) that specifically challenges the ability of AI Agent to comprehend complex interaction within long contexts. As illustrated in Table~\ref{tab.data_eg1}, the input contexts consist of 11 turns involving intricate interactions among three players and an NPC. C2A requires methods to accurately determine the character who will act in the next turn and the corresponding actions. To be noticed, actions in TRPGs can be simpify and classified as various pre-defined skills. GMs need to guide the players to operate correct skill checks in every games, which leads all intended actions of players are intuitively annotated by GMs and recorded in the game log. This induces the game logs can naturally involve labeled character intentions and C2A can leverage the game logs to construct the intention understanding tasks with accurate intention labels. 

%Complex Interaction based Character Intention estimation (CIE) 

\begin{figure*}[t]
    \centering
    \includegraphics[width=1.0\linewidth]{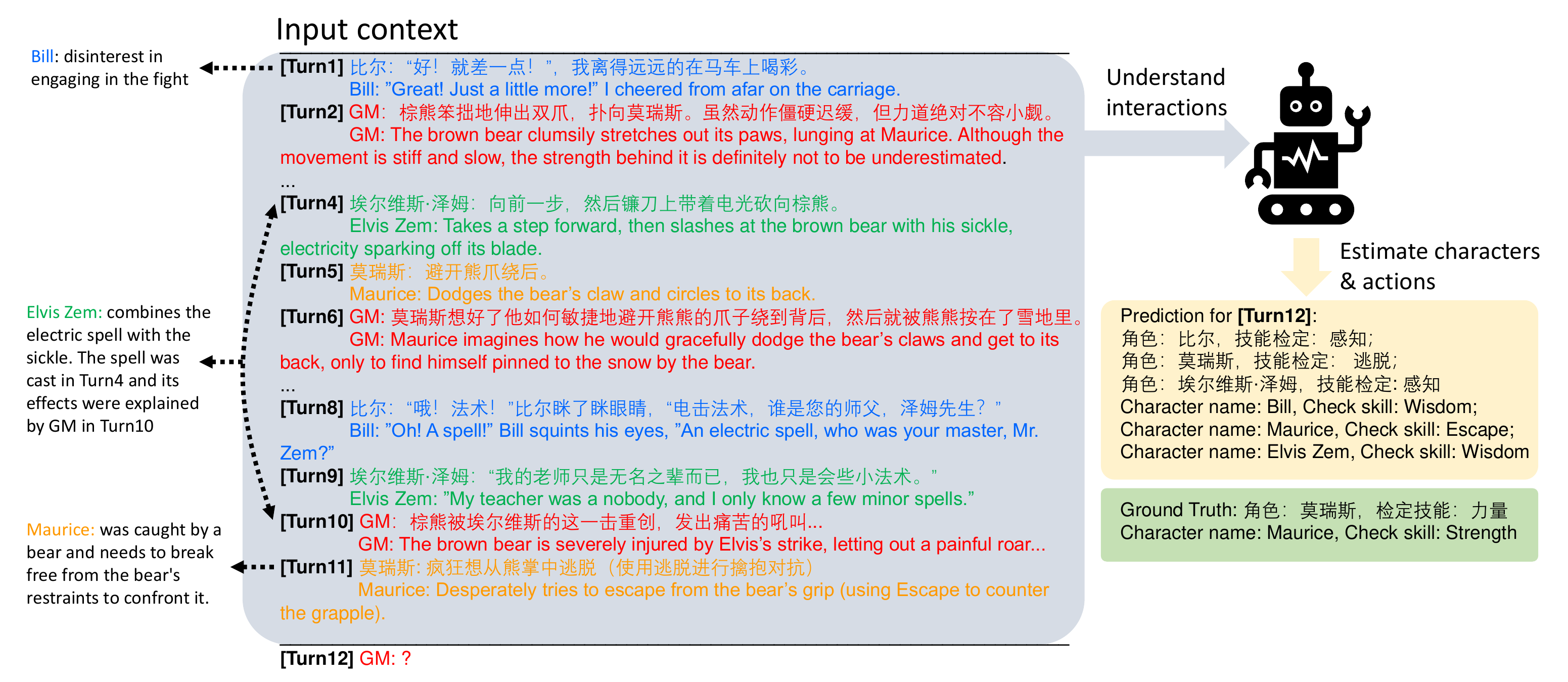}
    \caption{Example of MOE. In the given context, a scenario unfolds where three players find themselves facing a formidable brown bear in combat. Each character actively participates in the battle, except for Bill, who observes from the safety of a carriage. During the encounter, Zem casts a spell; however, it is important to note that the skill check for this particular spell has already been performed after Turn 4 and was explained by the DM in Turn 10. Consequently, the only character currently requiring a skill check is Maurice. Despite his intention to escape from the bear, the DND rule does not include a specific ``escape'' skill. In such a predicament, Maurice must utilize his strength to resist the bear's attempt to grapple him. As a result, the DM advises him to perform a strength check in adherence to the DND rule. Furthermore, we also present the predicted results from GPT-3.5 utilizing template prompts. The results demonstrate a lack of effective context comprehension and highlight the challenges in understanding complex interactions among agents.  }
    \label{fig:intro_moe}
\end{figure*}

We introduce a novel task, Multiple character and novel Object based interaction Estimation (MOE), which presents a challenge to comprehend complex interactions within long contexts. The input contexts, illustrated in Fig.~\ref{fig:intro_moe}, involve 11 turns encompassing intricate interactions among three players and an NPC. In MOE task, the primary objective is to accurately determine the character who will act in the next turn and identify the corresponding actions. It is important to note that actions in Tabletop Role-Playing Games (TRPGs) can be simplified and classified as various pre-defined skills. Game Masters (GMs) play a crucial role in guiding players to perform correct skill checks during gameplay, resulting in GMs intuitively annotating all intended actions, which are recorded in the game log. As a result, the game logs naturally contain labeled character intentions, enabling MOE to leverage this data to construct intention understanding tasks with accurate intention labels.

Moreover, there are two primary challenges that need to be addressed in MOE. Both challenges requires the methods to provide higher understanding ability to the input interactions. 
The first challenge revolves around comprehending the behaviors and intentions of multiple characters in complex scenarios. As depicted in Fig.~\ref{fig:intro_moe}, the current game scenario involves four characters: the brown bear, Bill, Elvis Zem, and Maurice. While all characters interact with one another, only one player intends to perform an action and needs to undergo a skill check in the upcoming turn. In the first turn, Bill expresses his disinterest in engaging in the fight. Subsequently, Zem combines the electric spell with the sickle. Notably, the spell was cast in Turn 4 and its effects were explained by the GM in Turn 10. Thus, the spell's execution has already taken place and should not be reevaluated after Turn 10.
%The first challenge involves understanding the behaviors and inferring the intentions of multiple characters. As in Tab.~\ref{tab.data_eg1}, four characters are involved in the current game scenario: the brown bear, Bill, Elvis Zem, and Maurice. While all characters have movements and interact with others, only one player intend to operate the action and needs to perform a skill check at the next turn. Bill hinted that he does not want to engaged in the fight as in the first turn. Zem adds the electric spell to the sickle. Zem's spell was cast in Turn 4, and its effects were explained by the GM in Turn 10. The spell has already been executed and should not be checked again after Turn 10.
The second challenge is understanding the game rules and aligning them with the characters' movements. In Fig.~\ref{fig:intro_moe}, Maurice intends to escape from the bear's attack. However, there is no specific `escape' operation in the skill checks defined by the DND rules. Instead, the bear utilizes its strength to grapple Maurice in the game, and Maurice must also check their strength to contest against the bear. To answer this skill check, methods need to comprehend the intentions and movements of characters and, based on the game rules, infer the appropriate check items for the current turn, akin to a real-human.

\subsection{Evaluation}
To provide a comprehensive assessment of context understanding in MOE, we evaluate the predicted character names and overall predictions separately. Specifically, we measure the average Character Precision (CP) and Character Recall (CR) for character names, as well as the average Skill Precision (SP) and Skill Recall (SR) for both character names and associated skills. Additionally, we compute the Character F-scores (CF) for character names and Skill F-score (SF) for both character names with associated skills.
\begin{align}
    \text{CP} & = \frac{1}{K} \sum_{i}^{K} t_c^i / p^i \\ 
    \text{CR} & = \frac{1}{K} \sum_{i}^{K} t_c^i / g^i \\ 
    \text{SP} & = \frac{1}{K} \sum_{i}^{K} t_s^i / p^i \\ 
    \text{SR} & = \frac{1}{K} \sum_{i}^{K} t_s^i / g^i \\ 
    \text{CF} & = 2 * \text{CP} * \text{CR} / (\text{CP} + \text{CR}) \\ 
    \text{SF} & = 2 * \text{SP} * \text{SR} / (\text{SP} + \text{SR}) 
\end{align}
where $i$ indicates the $i$-th sample, $t_c$ represent the number of correctly predicted character names, $t_s$ denote the number of correct predictions for both character names and associated skills, $p$ indicate the total number of predicted tuples, $g$ represent the number of answers in the ground truth, and $K$ represent the total number of evaluation data samples.

The metrics CP and CR are employed to evaluate the understanding of character intentions, focusing on the accuracy of predicting the characters about to take action. The proposed methods are required to provide correct character predictions, thereby achieving higher values for CP and CR. Then, to achieve higher SP and SR, the proposed methods must accurately comprehend both character intentions and the rules of the game. It is worth noting that if the model consistently predicts all characters as outputs, it may increase the recall but reduce the precision. Conversely, if the method tends to predict only one character, it may achieve higher precision but lower recall. To strike a balance between these factors, we employ F-scores as the final evaluation metrics in our experiments. The F-scores consider both precision and recall values, providing a comprehensive measure of the performance.

%CP and CR can reflect the understanding of the characters' intention, in which the proposed methods need to answer correct characters that about to initiate an action. Then, SP and SR further consider the game rule. To achieve a higher SP and SR, the proposed methods need to correctly understand both the intentions of character and the rules of games. Moreover, sometimes, if the model always predicts all characters as outputs, this may increase the value of recall but reduce the precision. Meanwhile, if the method tends to predict only one character, though may with higher precision, this induce lower recall values. To balance the two factors, we use F-scores as the final metrics in our experiments, which consider the values of both precision and recall.

\begin{figure*}[htbp!]
    \centering
    \includegraphics[width=0.85\linewidth]{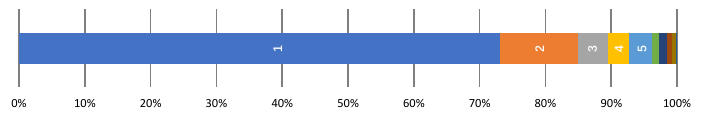}
    \caption{Distribution of character number in MOE labels. }
    \label{fig.sc_cnt_sta}
\end{figure*}

\begin{figure*}[htbp!]
    \centering
    \includegraphics[width=0.85\linewidth]{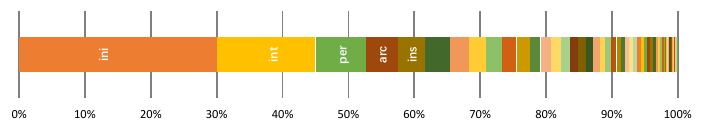}
    \caption{Distribution of skill names in MOE labels of the contexts within DND rule. initiative (ini), intelligence (int), perception (per), arcana (arc), insight (ins). }
    \label{fig.sc_sta}
\end{figure*}

\subsection{Skill Check Annotation}
In Tabletop Role-Playing Games (TRPGs), skill checks can directly indicate the players' intentions in the game, which play a crucial role in determining the success or failure of character actions. When a player wishes to have their character perform an action involving a skill, such as combat, persuasion, or searching for clues, the game models or rules provide a difficulty level or target number for the action. This difficulty level represents the challenge or desired level of success for the action. The Game Master (GM) assumes the responsibility of judging and guiding the player in performing the skill check based on the character's proficiency associated with the action. The GM then rolls a dice to determine the outcome.
In our task, we annotate the skill checks performed by players' characters during the games based on the semantic or recorded results provided by the GM. It is important to note that some skill checks are not predictable based solely on previous interactions. For example, in Call of Cthulhu (COC) games, perception checks may be prompted by the GM when players enter specific locations. These checks are closely tied to the game models and cannot be predicted in advance. Additionally, certain checks require additional calculations based on character attributes or cards, such as determining damage based on hit points or factoring in armor class to reduce damage. These calculations cannot be inferred solely from the game logs and we also remove these check in MOE.
Thus, we have excluded any checks that are unpredictable and included only those check items that can be inferred from the game logs. For example, the COC logs contain 61 check items (including skills and attributes) that can be verified, while the DND logs contain 25 such items. Further details regarding the check items will be provided in the supplementary material.

\subsection{Context Excerpt}
Following the labeling of check items in the game logs, we proceed to excerpt relevant contexts associated with each check. Our selection of excerpted contexts is guided by three key principles to ensure the inclusion of long and diverse interactions.
First, we ensure that the excerpted contexts encompass complete events within the game, such as the entire process of encountering enemies or the detailed information and clues leading up to the exploration of certain locations. This ensures that the extracted contexts provide a comprehensive understanding of the events.
Second, we require that the excerpted contexts involve at least two characters who are present in the current scenarios of the game. This criterion allows for the examination of interactions between multiple characters, providing a more complex context for analysis.
Lastly, we ensure that at least one character within the excerpted contexts has a skill check that can be predicted. This principle guarantees that the selected contexts contain situations where skill checks can be inferred based on the information available up to the last turn.
By adhering to these principles, we ensure that the contexts support the understanding of the complex interactions and enable the inference of characters' intentions in subsequent turns.

%After labeling check items in the game logs, we excerpt related contexts to the check. To reserve long and diverse interactions, we excerpt contexts follows three principles: 1. The excerpted contexts should involve a complete event (e.g., the complete process of encountering enemies, the detailed information and clues before entering some places). 2. There should be at least two characters in current scenarios of the game. 3. At least one character should have predictable skill check. These principles ensure that the contexts in MLG support the understanding of multi-character interactions and can be used to infer skill check after the last turn. 

\subsection{Statistical Analysis}
We present the statistical results of answers in MOE in Tab.~\ref{tab.dataset_sta}. In total, we have extracted and labeled 1,003 sets of contexts and corresponding skill checks, which serve as the input context and ground truth for our task. The average number of turns in our dataset is 32.12, indicating its complexity compared to previous works that primarily focused on single-turn responses. %To facilitate evaluation with various Large Language Models (LLMs), we have also translated the logs into English, and the details of the translated version can be found in the supplementary material.
Furthermore, we provide the distributions of skill check labels of the Dungeons and Dragons (DND) logs in the MOE task, as illustrated in Fig.~\ref{fig.sc_sta} and Fig.~\ref{fig.sc_cnt_sta}. The number of characters involved in skill checks varies from 1 to 11, with an average of 1.696 characters per skill check. This reflects the complexity of multi-character interactions within our dataset, which close to the real-human communication in the games. Additionally, the items for skill checks exhibit diversity, highlighting the varied interactions between players. Both sets of statistical results underscore the value of our task as a comprehensive testbed for understanding complex interactions in TRPGs.

%We present some statistical results in Tab.~\ref{tab.dataset_sta}. In overall, we excerpt and label 1,003 sets of contexts and corresponding checks, which formulates as the input context and the ground truth of our QA task in the next section. The average turns in our dataset is 32.12, which is more complex than only 1 turn response in previous works. To better evaluate with various LLMs, we also translate the logs to English and the details for translated version is in supplementary.
%Moreover, the distributions of labels in DND logs of CSA are illustrated as in Fig.~\ref{fig.sc_sta} and Fig.~\ref{fig.sc_cnt_sta}. The number of characters for skill checks are variable from 1 to 11. The average number of characters is 1.696. This reflects the property of multi-character interaction in our dataset. Besides, items for skill check are diverse. This reveal the diversity of the interactions between players. Both statistic results can show the value of our dataset as a testbed for understanding multi-character interactions and grounded language. 

\section{Dataset}
To support our MOE task with more grounded and complex data, we have collect a new dataset. It is sourced from a Chinese TRPG forum\footnote{www.goddessfantasy.net}. This forum hosts a wide array of game records uploaded by users, spanning different rule systems e.g., DND, COC, PF, SW, etc. Unlike play-by-post forums~\cite{callison-burch-etal-2022-dungeons}, where players interact by writing and posting responses, the game logs in this forum are compiled and summarized by the Game Masters (GMs) after the game sessions\footnote{Most users whose logs are used in our study have agreed and provided their informed consent. We are trying to contact and communicate all users to be informed and agree with the participation of the research.}. Besides, in play-by-post scenarios, interactions between players are not immediate, and the feedback from the next player may not appear until several days or even weeks later. In contrast, the majority of game logs in our dataset are derived from instant messaging platforms, including voice and text communication. This characteristic allows for the capture of abundant immediate player responses, closely mirroring daily conversations with grounded language interactions. Consequently, our dataset provides more grounded semantics within real-time communication, making it conducive for exploring AI agents.

Statistically, our dataset comprises 95 sets of records from different games with various rule systems. It encompasses a total of 647,480 Chinese words, as indicated in Tab.~\ref{tab.dataset_sta}. In summary, our dataset not only surpasses previous works in terms of data diversity, groundedness, and complexity but also matches or exceeds their scale.

\begin{table}[t]
	\begin{center}
	\resizebox{0.55\columnwidth}{!}
    {
        \begin{tabular}{|l|l|l|}
        \hline
        Dataset    & \#words        & rules         \\ \hline \hline 
        DDD Corpus~\cite{louis2018deep} & $\sim$4,430,000                      & DND           \\ \hline
        PBP~\cite{callison-burch-etal-2022-dungeons}       & 58,187,526                          & DND           \\ \hline
        GANDALF~\cite{gandalf}    & $\sim$47,000                & DND           \\ \hline
        Ours        & 647,480                  & DND,COC,PF,SW \\ \hline
        \end{tabular}
    }
    \end{center}
    \caption{Dataset statistic. Our dataset exhibits a comparable scale to previous works, while also encompassing a higher diversity of game rules. }
    \label{tab.dataset_sta}
\end{table}

\section{Think Before Speak prompting method}
We propose a three-step agent generation benchmark called ``Think Before Speak'' (TBS), which aims to guide Large Language Models (LLMs) in comprehending complex and lengthy contexts of interactions more accurately. Unlike simple template-based prompting approaches~\cite{llm_survey,zhao2023survey,huang2022towards}, our method takes into consideration the specific properties of Tabletop Role-Playing Games (TRPGs) and incorporates the principles of Chain of Thought (CoT)~\cite{wei2022chain,zero-shot-cot} in its prompting design.
In the generated check item, the answers consist of character names and corresponding skill names. However, directly expecting the models to produce accurate character and skill names is a challenging task. Using a single-step template prompting approach may result in LLMs generating characters that do not exist in the given contexts, characters with no relevant actions, mismatches between characters and their associated skills, or skills that are not defined within the game rules.
To address these challenges, our method guides LLMs through a three-step process. Firstly, the models are prompted to identify the characters present in the current game scenarios. Then, they are encouraged to consider the intentions of the characters and list those who are likely to take action or are engaged in ongoing movements. Finally, we provide the models with a comprehensive set of possible skills derived from the game rules, allowing them to select the most appropriate character-skill combinations that the GM may ask the players to check. This gradual guidance facilitates more accurate and context-aware responses from the LLMs.

%We introduce a three-step virtual GM generation benchmark named think before speak (TBS), which can gradually guide LLMs to more accurately understand the complex and long contexts. 
%Rather than straightforward template prompting~\cite{llm_survey,zero-shot-cot}, ours consider the properties of TRPG games and leverage CoT in prompting design. In the generated check item, the answers of skill check contains two components, which are the characters' names and corresponding skill names. Directly requiring the models to produce correct names and skills is challenging. With the single-step template prompting, LLMs may produce characters that not existed in the contexts, produce characters without any movements, produce mismatch between characters and corresponding skills, or produce non-existed skills in the rules. In our work, we guide LLMs first to figure out which characters are in current game scenarios, and then think about the intentions and list characters that may act next or perform some ongoing movements. Finally, we given all the possible skills from the rules and let the models to select most possible characters and corresponding skills that DM may ask the players to check. 

Specifically, in the first step of our prompting approach, we guide the language models by providing a prompt such as ``Based on the TRPG game record provided above, identify the characters or NPCs that exist in the current scenarios.'' This prompts the language model to recognize and understand the characters present in the given contexts.
In the second step, we prompt the language models with a question like ``Which character or NPC is expected to carry out activities next?'' This encourages the models to delve deeper into the semantics of the contexts and infer the intentions of the characters.
For the final step, we provide LLMs with all possible skills defined in the TRPG rules and guide them to generate character names that correspond to the potential skill checks. Our prompts for this step include phrases such as ``What skills are required for the mentioned characters to carry out their respective activities?''
Furthermore, to facilitate comparison with other benchmarks, we extend the TBS approach to also generate utterances to simulate a real-human GM. Given the predictions from the TBS model, LLMs are required to generate responses in the tone and style of a GM. We achieve this by using prompts such as  ``As a game master for a TRPG game, generate responses based on the provided character names and the corresponding skills.''

\section{Experimental Results}
In this section, we provide a detailed discussion, comprehensive evaluation, and analysis of our benchmark.

\textbf{Baseline Methods:} As our baseline, we employ LLMs with template prompting, which have been utilized in previous studies. We specifically adapt two popular LLMs, which are GPT-3.5 and GPT-4~\cite{brown2020language}. By incorporating different language models, we can thoroughly assess the performance of our prompting benchmark. Furthermore, recent researches~\cite{wei2022chain,zero-shot-cot,cot_self_consistency} have demonstrated the efficacy of Chain-of-Thought (CoT) methods in improving understanding capabilities. To compare with this approach, we include the zero-shot CoT (zcot) method~\cite{zero-shot-cot} in our evaluation.

Additionally, to demonstrate the ability to infer check items, we introduce a statistical predictor for check items. Given the predicted characters, we select the skills with the highest probability based on the statistical distribution observed in our dataset. This statistical predictor serves as a lower bound for generating check items and also reveals the impact of any biases present in our dataset.

%\textbf{Baselines: } We consider LLMs with simple template prompting as our baseline, which has been introduced and presented in previous methods. We adapt three popular LLMs, which are GPT-3.5, GPT-4, LLAMA, respectively. Different language models can comprehensively reflect the performances of our prompting benchmark. Meanwhile, many recent works show the effectiveness of Chain-of-Thought (CoT) methods in improving understanding ability. We also adapt zero-shot CoT (zcot) method~\cite{zero-shot-cot} in comparison. Moreover, to show the ability of inferring check items, we further introduce a statistical predictor for check items. Given predicted characters, we select the skills with the highest probability based on the statistical distribution of the dataset. This statistical predictor reflect the lowest boundary of generating check items and also reveal the influence of bias in our dataset.

\textbf{Evaluations:} To evaluate the effects of MOE and TBS frameworks on interaction understanding, we introduce the concept of a virtual Game Master (GM) in TRPGs. The virtual GM serves as a simulation of a real-human GM, possessing the ability to comprehend interactions, infer intentions, interact with players, and provide guidance for their actions. This role fulfills the criteria of our requirements for the agents that enable to understand complex interactions. By incorporating the virtual GM, we create a platform to assess the agents' understanding of complex interactions and their ability to navigate diverse scenarios.
%In order to assess the impact of MOE and TBS on the understanding of interaction, we propose to generate a virtual GM participate in the games. The virtual GM needs to simulate the response of real-human GM, who comprehend interactions, estimate intentions, interact with players and guide them to act. This role is just satisfy the requirements of our world model for the agents, which can indicate the agents whether they can successfully cooperate with others or interact with open objects. 
In detail, we generate GM utterances using both ground truth information from C2A and predictions from TBS. The generation process follows the methodology outlined in~\cite{zhu2023fireball,llm_survey}, which leverages LLMs, template prompts, and additional inputs for characters and skills.

Rather than relying on metrics based on captioning in previous works~\cite{gandalf,zhu2023fireball}, we employ subjective evaluation conducted by real-human players. Given the diversity of descriptions in grounded language, there is no definitive ground truth for evaluating the responses of GMs. Subjective evaluation provides more valuable insights into the degree of realism in the generated utterances. Following~\cite{gandalf,sub_eval1,sub_eval2,sub_eval3,liang2022seeg}, we invite volunteers to score the responses based on three factors: naturalness, groundedness, and factual correctness. Naturalness assesses the extent to which the generated responses resemble human-like language. Groundedness measures the degree to which the responses effectively employ grounded language similar to everyday communication. Lastly, factual correctness evaluates whether there are any factual errors or inconsistencies with the given contexts.

%\textbf{Virtual GM Evaluation: } To evaluate the effect of CSA and TBS to virtual GM generation, we further generate GM's utterance by given the ground truth in CSA and predictions from TBS. The generation method following~\cite{zhu2023fireball,llm_survey}, which utilize LLMs, template prompts, and additional inputs for characters and skills in ours. Moreover, rather than compare some metrics based on captioning, we adapt subjective evaluation by real-human players. Due to the diversity of descriptions in grounded language, there is no proper ground truth for the responses of GMs. The subjective evaluation can provide more valuable results for measuring the vivid degree of utterances. Following~\cite{gandalf,sub_eval1,sub_eval2,sub_eval3,liang2022seeg}, we ask the invited volunteers to score the responses in three factors, which are naturalness, groundness and factually correctness. The naturalness reflects how much the produced responses are similar to real-human. Groundness reflects what degree the generated responses flexibly use grounded language similar to the daily communication. Finally the factually correctness reflects whether there are some factual errors or inconsistency with the contexts. 

% \usepackage{multirow}
\begin{table}[]
	\begin{center}
	\resizebox{0.88\columnwidth}{!}
            {
\begin{tabular}{|l|lllc|}
\hline
\multicolumn{1}{|c|}{\multirow{3}{*}{Prompting Method}} & \multicolumn{4}{c|}{LLMs}                                                                             \\ \cline{2-5} 
\multicolumn{1}{|c|}{}                                  & \multicolumn{2}{c|}{GPT-3.5}                      & \multicolumn{2}{c|}{GPT-4}                        \\ \cline{2-5} 
\multicolumn{1}{|c|}{}                                  & \multicolumn{1}{c|}{CF} & \multicolumn{1}{c|}{SF} & \multicolumn{1}{c|}{CF} & \multicolumn{1}{c|}{SF} \\ \hline \hline 
template prompt                                         & \multicolumn{1}{c|}{42.02}   & \multicolumn{1}{c|}{15.30}   & \multicolumn{1}{c|}{43.21}  &  15.93           \\ \hline
template prompt + zcot                                  & \multicolumn{1}{c|}{39.28}   & \multicolumn{1}{c|}{14.46}   & \multicolumn{1}{c|}{42.45}  &  16.25           \\ \hline \hline 
char prompt + skill prompt                              & \multicolumn{1}{c|}{50.43}   & \multicolumn{1}{c|}{14.78}   & \multicolumn{1}{c|}{53.55}  &  16.79           \\ \hline
pre-char prompt + char prompt + statistic predictor     & \multicolumn{1}{c|}{53.32}   & \multicolumn{1}{c|}{5.03}    & \multicolumn{1}{c|}{57.94}  &  5.03            \\ \hline
pre-char prompt + char prompt + skill prompt + zcot     & \multicolumn{1}{c|}{50.50}   & \multicolumn{1}{c|}{12.88}   & \multicolumn{1}{c|}{53.45}  &  17.39           \\ \hline
pre-char prompt + char prompt + skill prompt            & \multicolumn{1}{c|}{53.32}   & \multicolumn{1}{c|}{15.91}   & \multicolumn{1}{c|}{57.94}  &  20.02           \\ \hline
\end{tabular}
             }
        \end{center}
        \caption{Comparison of different prompting methods and LLMs. Results prove that our task is solvable but requires higher understanding ability for grounded and complex semantics. }
        \label{tab:p_r_metric}
\end{table}

\subsection{Objective Evaluation}

\textbf{Comparison of Prompting Methods:} We conduct a comparison between our proposed method and different prompting approaches. The results, as shown in Tab.~\ref{tab:p_r_metric}, reveal the effectiveness of our step-wise prompting approach compared to baselines such as zero-shot CoT and the statistical predictor.
The experimental results demonstrate that each step in our prompting process contributes significantly, leading to improved F-score for both characters and skills. This highlights the enhanced understanding capability of LLMs in comprehending the given contexts. Furthermore, due to the distribution bias present in our dataset, the statistical predictor proves to be useful, albeit with considerably lower performance compared to our proposed method and other prompting methods. This reveal the lower performance boundary in predicting skill labels.

Furthermore, in line with previous studies~\cite{zero-shot-cot,dong2022survey,wei2022chain}, the incorporation of zero-shot CoT has demonstrated improvements in the performance of LLMs across various tasks. However, when applied to the MOE task, the observed enhancements are not as substantial. Since MOE involves more grounded semantics and complex interactions, it presents a challenging scenario for existing prompting methods and remains an unsolved problem that requires further investigation.

%\textbf{Comparison of different prompting method: } We first compare our method with different prompting methods. As shown in Tab~\ref{tab:p_r_metric}, we experiment prompting methods with different steps and compare with baselines with zero-shot CoT and statistical predictor. The experimental results show that all step in our prompting is valuable, improve the precision and recalls in both characters and skills, and boost the understanding ability for LLMs to current contexts. Besides, since the distribution bias of our LGL, statistical predictor is also useful but its performances are far lower than ours and other prompting methods. This reveal a lower boundary of the performances of predicting skill labels. Finally, as presented in previous works~\cite{zero-shot-cot}, zero-shot CoT can improve the understanding ability of LLMs and lead the models to think gradually. This is also useful in our benchmark and introduce additional improvements in results. 

%\textbf{Comparison of different languages: }

\textbf{Comparison of different language models: } We further investigate the impact of different LLMs on the performance of our prompting methods. With advancements in LLM, the overall understanding and reasoning capabilities have significantly improved. As depicted in Tab.~\ref{tab:p_r_metric}, employing more advanced language models leads to higher performance in MOE task. In addition to the effectiveness of the prompting methods, the enhancements in LLMs themselves are also beneficial in comprehending the intricacies of complex and grounded interactions. The experimental results reveal that our task is solvable, yet there remains ample room for further exploration and improvement.

%\textbf{Comparison of different language models: } We also experiment the prompting methods with different LLMs. Since the developments from LLMs, the understanding and reasoning ability have been boosted. As in Tab.~\ref{tab:p_r_metric}, using better language model achieve higher performances in CSA. More than prompting methods, the improvements of LLMs are also useful in understanding the complex and grounded semantics. The experimental results reveal that our task is solvable but still have a large space to explore. 

\subsection{Subjective Evaluation}
\begin{figure*}[htbp!]
    \centering
    \includegraphics[width=0.8\linewidth]{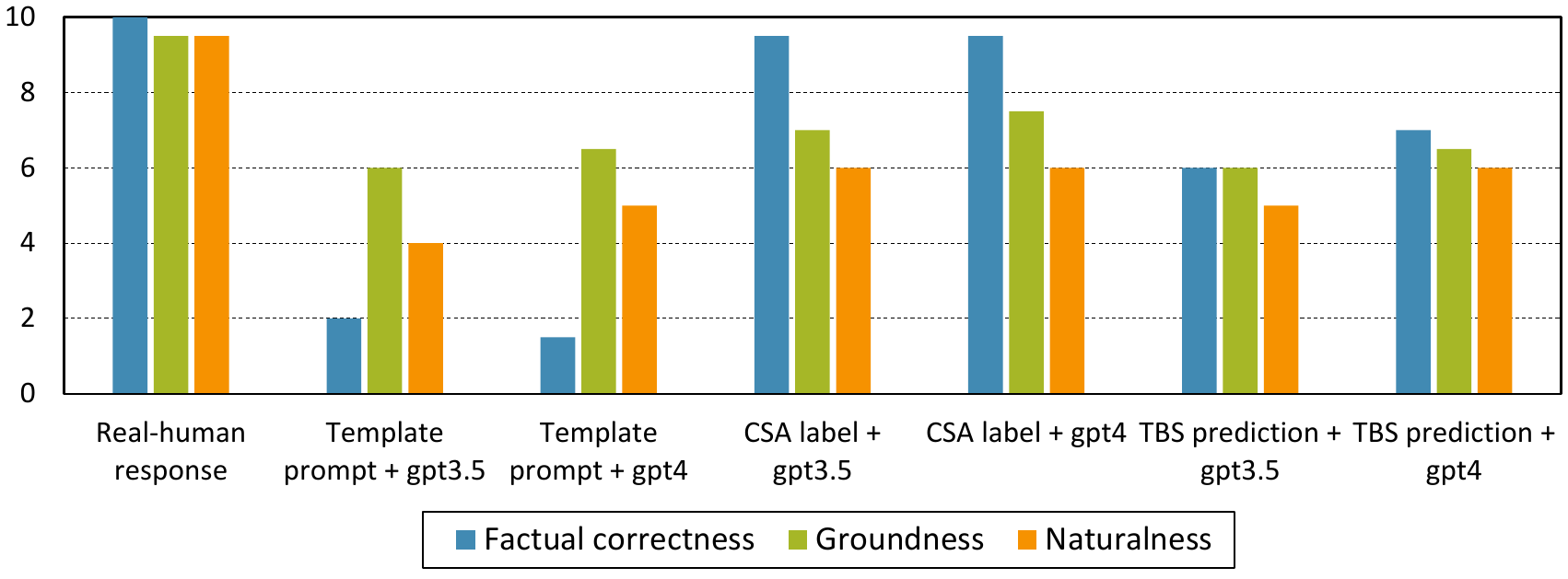}
    \caption{Subjective evaluation by volunteers. With MOE labels or predictions from our method, LLMs generate better responses that close to the real-human in all three evaluating factors. }
    \label{fig:subjective}
\end{figure*}

We conducted a subjective evaluation by recruiting real-human players of TRPG as volunteers and collecting their responses through questionnaires. The average scores in different factors, which are naturalness, groundedness, and factual correctness, were computed following established guidelines~\cite{gandalf,sub_eval1,sub_eval2,sub_eval3,liang2022seeg}. The statistical results are presented in Fig.~\ref{fig:subjective}. Notably, methods that take into account the predictions or ground truth of MOE demonstrate higher performance across all evaluation factors. Generally, methods utilizing MOE labels outperform those using predicted labels. Moreover, when considering MOE predictions, the methods achieve superior performance in generating virtual GM responses. This observation confirms that a higher understanding ability for complex semantics leads to more vivid and human-like responses from the agents. Additionally, it underscores the strong correlation between MOE performance and virtual GM performance, highlighting the importance of MOE in the pursuit of improved agent generation.

%We conducted a subjective evaluation by enlisting real-human players of TRPG as volunteers and collecting their responses through questionnaires. The average scores in different factors, which are naturalness, groundedness, and factual correctness, were computed following established guidelines~\cite{gandalf,sub_eval1,sub_eval2,sub_eval3,liang2022seeg}. The statistical results are presented in Figure~\ref{fig:subjective}. To be noticed, methods considering the predictions or the ground truth of CSA show higher performances in all evaluating factors. Generally, methods with CSA labels outperform methods with predicted labels. Meanwhile, considering CSA predictions, methods also achieve higher performances in generating virtual GM responses. This prove that with higher understanding abbility for complex semantics, the agents can be more vivid and close to the real human. Meanwhile, this also reveal the high relevance of our CSA performances and virtual GM's performance. This reflect the significance of our CSA in pursuing better agent generation. 

Besides, our prompting method demonstrates superior performance in all evaluated factors. Specifically, our method exhibits significant improvements in factual correctness compared to the baseline methods. Furthermore, in terms of groundedness and naturalness, our method showcases comparable or even better performance than other methods. These results indicate that our method achieves enhanced understanding ability and is capable of generating improved utterances as GM descriptions. However, there is still ample room for improvement in terms of groundness and naturalness. The generated utterances may occasionally be overly verbose and lack the same level of vividness as those produced by real humans. This performance gap motivates further exploration of more effective methods for constructing advanced AI agents.

%\subsection{Subjective Evaluation}
%We invite real-human players of TRPG as volunteers and collect the questionnaires from them. We compute the average score in different factors, which are naturalness, groundness, and factually correctness, following~\cite{gandalf,sub_eval1,sub_eval2,sub_eval3,liang2022seeg}. The statistical results are presented as in Fig.~\ref{fig:subjective}. Our method show better performances in all factors. Specifically in factually correctness, our method show significant improvements than baseline methods. Meanwhile, for groundness and naturalness, our method also show comparable or better performances than other methods. The results indicate that our method achieve better understanding ability and also enable to produce better utterances as descriptions for GM. However, there are still large space to improve in groundness and naturalness, the generated utterance may tend to be wordy and still not as vivid as real-human. This performance gap also further encourages us to explore more effective methods to construct better AI agents. 

\section{Conclusion}
This paper proposes a new dataset, task, and benchmark to enhance the understanding ability of AI agents in dealing with complex interactions with multiple characters. The existing works in this field have limitations, particularly their reliance on forum-based data collections and do not consider complex and grounded semantics in the real-time communications. To overcome these limitations, we formalize a new task named Multiple character and Open instances based interaction Estimation (MOE), providing a testbed for the understanding ability of the agents and leading further improvements in agents' factual correctness. We also introduce a dataset to support MOE task, which is derived from real-time game logs in tabletop role-playing games (TRPGs) and provides a richer and more complex context capable of supporting MOE tasks.
Additionally, we introduce a prompting benchmark designed specifically to refine the interaction capabilities of AI agents in TRPGs. This benchmark focuses on understanding complex interactions and generating vibrant game master utterances. The three-stage generation process, which includes game check and GM utterance generation, has been evaluated both objectively and subjectively. The results clearly indicate that this approach significantly enhances the quality of AI responses within the TRPG context. We hope that this work will serve as inspiration for the AI community to further explore and enhance their understanding of complex grounded interactions and advance the interaction ability of AI agents.

%In this paper, we propose a new dataset, task and benchmark to enhance the understanding ability of AI agents for long and grounded semantics in Tabletop Role-Playing Games (TRPGs). Due to the limitations of current works, particularly their reliance on forum-based data collections and end-to-end utterance generation, we introduces the novel Long-context Grounded-language TRPG Logs dataset (LGL). This dataset, derived from real-time game logs, provides a richer, more complex context, which can support CSA. 
%We further introduce a prompting benchmark, designed to refine the generation capabilities of AI agents in TRPGs. It focuses on understanding complex interactions and generating vibrant game master utterances. The three-stage generation process, consisting of game check and GM utterance generation, has been evaluated both objectively and subjectively. The results indicate that this approach can significantly improve the quality of AI responses in a TRPG context. We hope this work will inspire the AI community to further explore and enhance the understanding of complex grounded semantics and the generation of AI agents.

\section{Limitations and Social Impacts}
While the use of an AI agent in a tabletop role-playing game (TRPG) could revolutionize the way these games are played, providing consistent and unbiased decisions, there are potential limitations and social impacts to consider. One key limitation is the AI's ability to simulate human creativity, empathy, and adaptability, which are all fundamental to the role of a game master. For instance, the AI may not fully comprehend nuanced player interactions or adapt the game based on the players' emotional state. Additionally, there could be social implications, such as the potential reduction in human interaction and shared storytelling, which are often crucial elements of TRPGs. For players, part of the joy of a TRPG is the shared human experience, the unpredictable responses, and the subtle non-verbal cues, which an AI might not replicate. The introduction of an AI game master could also result in job loss in professional game-mastering circles. Despite the AI's potential to provide a consistent and more accessible gaming experience, these human and social elements may be irreplaceable in a TRPG context.

%\bibliographystyle{neurips_data_2023}
%\bibliography{reference}
{
		\small
		\bibliographystyle{ieee_fullname}
		\bibliography{reference}
}

\end{document}